\documentclass[final]{cvpr}

\usepackage{epsfig}
\usepackage{graphicx}
\usepackage{amsmath}
\usepackage{amssymb}
\usepackage{booktabs}
\usepackage{dsfont, eucal}

\usepackage[margin=6pt,font=small,labelfont=bf,labelsep=endash,tableposition=top]{caption}

\usepackage{xcolor}		%
\usepackage[colorlinks = true,
            linkcolor = purple,
            urlcolor  = blue,
            citecolor = cyan,
            anchorcolor = black]{hyperref}	%

\def\SexyName{{DyCo3D}\xspace}

\def\AcceptedNotice{{\it Appearing in IEEE/CVF Conf. Computer Vision and Pattern Recognition (CVPR), 2021. Content may change prior to final publication.}}

\begin{document}

\title{\SexyName: Robust Instance Segmentation of 3D Point Clouds
\\ 
through Dynamic Convolution\thanks{\AcceptedNotice}}

\author{
Tong He,  ~ ~ ~ Chunhua Shen\thanks{Corresponding author (e-mail: {\tt chunhua@icloud.com}).}, ~ ~ ~ Anton van den Hengel
\\[0.2cm]
The University of Adelaide, Australia
}

\maketitle
\begin{abstract}
	
	Previous top-performing approaches for point cloud instance segmentation involve a bottom-up strategy, which often includes inefficient operations or complex pipelines, such as grouping over-segmented components, introducing additional steps for refining, or designing complicated loss functions. 
	The inevitable variation in the instance scales can lead bottom-up methods to become particularly sensitive to hyper-parameter values. 
	To this end, we propose instead a dynamic, proposal-free, data-driven approach that generates the appropriate convolution kernels to apply in response to the nature of the instances.
	To make the kernels discriminative, we explore a large context by gathering homogeneous points that share identical semantic categories and have close votes for the geometric centroids. Instances are then decoded by several simple convolutional layers. 
	Due to the limited receptive field introduced by the sparse convolution, a small light-weight transformer is also devised to capture the long-range dependencies and high-level interactions among point samples. 
	The proposed method achieves promising results on both ScanetNetV2 and S3DIS, and this performance is robust to the particular hyper-parameter values chosen. It also improves inference speed by more than 25\% over the current state-of-the-art. Code is available at: 
	\rm 
    \url{https://git.io/DyCo3D}
\end{abstract}

\section{Introduction}
\begin{figure}[htbp]
	\centering
	{
		\includegraphics[width=8.5cm]{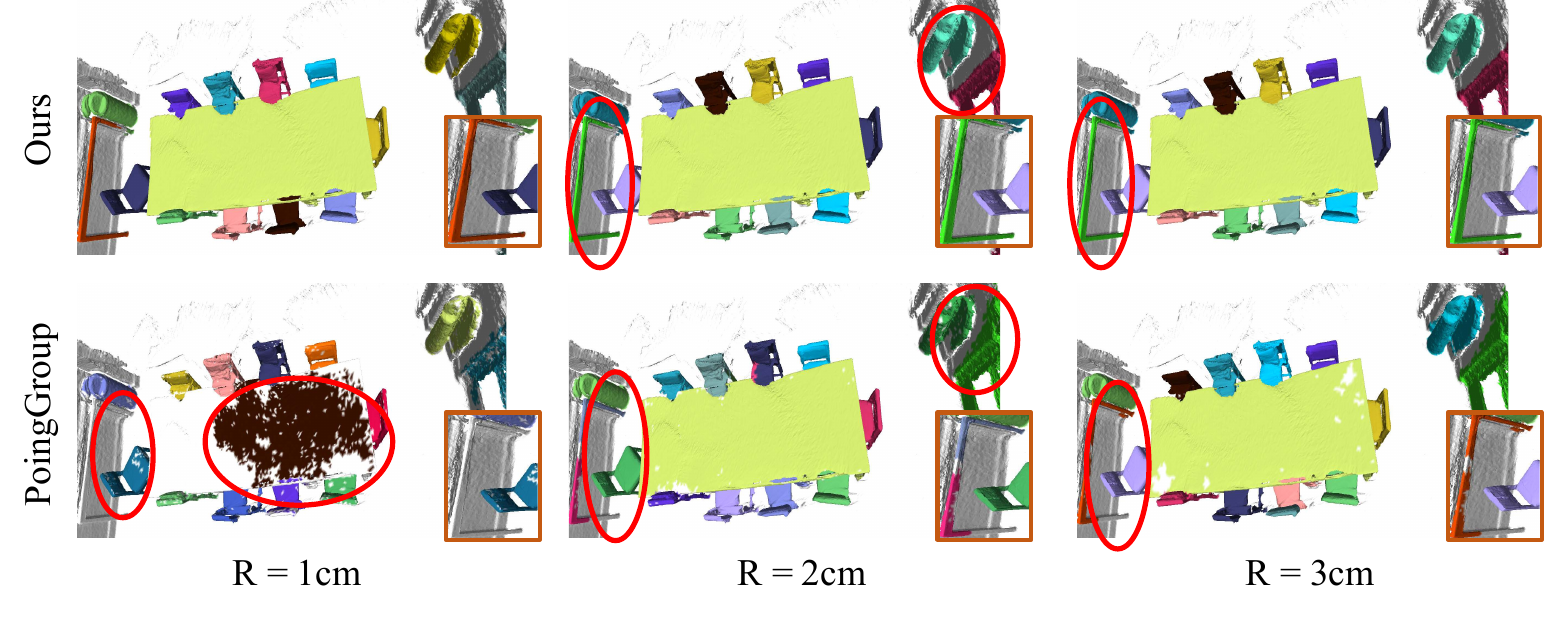}
	}
	\caption{A comparison of the instance segmentation results achieved using \SexyName, and PointGroup \cite{jiang2020pointgroup}. Our method shows better robustness and generalization to varying hyper-parameter values. Different instances are shown with random colors, and red ellipses highlight specific over-segmentation errors.  Best viewed in color.
	}
	\label{fig:main}
\end{figure}
Instance segmentation is significantly more challenging than semantic segmentation because it requires identifying every individual instance of a class of objects, and the visible extent of each. The information recovered has proven invaluable for scene understanding, however.
With the increasing applications of 3D sensors (such as LiDAR and laser scanners), point clouds have become an important modality in scene understanding.  
Although significant advances have been made in instance segmentation in the image domain \cite{he2018maskrcnn, chen2020cascade, chen2020blendmask, tian2020conditional}, instance segmentation of 3D point clouds has proven far more challenging. This is partly due to the inherent irregularity and sparsity of the data, but also due to the diversity of the scene. By way of example, Mask R-CNN \cite{he2018maskrcnn}, which has shown great success when applied to 2D images, performs poorly when applied in 3D \cite{hou20193dsis}. 

Many previous top-performing approaches for point cloud instance segmentation adopt a bottom-up strategy, involving heuristic grouping algorithms or complex post-processing steps. 
3D-MPA \cite{Engelmann20CVPR}, for example, extracts proposals from the predicted instance centroids. Instances are then generated by aggregating proposal-wise embeddings.
PointGroup \cite{jiang2020pointgroup} generates instances proposals by gradually merging neighbouring points that share the same category label. Both original and centroid-shifted points are explored with a manually specified search radius. A separate model (labelled ScoreNet) is used to estimate the objectness of the proposals. 
Both methods achieved promising performance on ScanNetV2 \cite{dai2017scannet} and S3DIS \cite{armeni2016s3dis} benchmarks. However, these bottom-up methods often suffer from several drawbacks: (1) the performance is sensitive to values of the pre-defined hyper-parameters, which require manual tuning. In PointGroup \cite{jiang2020pointgroup}, modifying the clustering radius from 3cm to 2cm causes mAP to drop by more than 6\%, illustrating the method's limited robustness and generalization ability. These results are presented in Fig. \ref{fig:main}. (2) they incorporate either complex post-processing steps or training pipelines, rendering them unsuitable for real-time applications such as robotics and driverless cars. For example, 3D-MPA \cite{Engelmann20CVPR} needs an extra 10-layer graph network and a clustering post-processing step to yield its final instance segmentation masks. (3) they are heavily reliant on the quality of the proposals, which limits their robustness and can lead to joint/fragmented instances in practice. 

In this paper, we propose a novel pipeline tailored to 3D point cloud instance segmentation using dynamic convolution, that we label \SexyName. Our approach addresses the task with only a few convolution layers, for which the filters are generated on the fly, conditioned on the category and position of the instance to be decoded. To empower the filters to distinguish different instances, we propose to encode category-specific context by deploying a light-weight sub-network to explore homogenous points that have close votes for instance centroids and share the semantic labels. 
Instance masks can be decoded in parallel by convolving the generated class-specific filters with the position embedded features. 
Compared with bottom-up approaches \cite{jiang2020pointgroup, Engelmann20CVPR, wang2019asis, wang2018sgpn, he2020eccvmemory} that are sensitive to the values of numerous hyper-parameters, our approach demonstrates superiority on both effectiveness and efficiency. Qualitative results illustrating this fact are presented in Fig. \ref{fig:main}.

Besides, as has been proved in the 2D image domain, a large receptive field and rich context information are critical to the success of instance segmentation \cite{cheng2020panoptic}. To address the problem in the 3D point cloud, we propose to introduce a small transformer \cite{Vaswani2017attention} to capture a long-range dependency and build high-level interactions among different regions.

Our main contributions are summarised as 
follows. 
\begin{itemize}
\itemsep 0cm
	
	\item A novel method for 3D point cloud instance segmentation based on dynamic convolution that outperforms previous methods in both efficiency and effectiveness.

	\item A proposal-free instance segmentation approach that is more robust than bottom-up strategies.

	\item A light-weight transformer that enlarges the receptive field and captures non-local dependencies.  

	\item Comprehensive experiments demonstrating that the proposed method achieves state-of-the-art results, with improved robustness, and an inference speed superior to that of its comparators.
\end{itemize}

\section{Related Work}
\textbf{Deep Learning for 3D Point Cloud.} In contrast to the image domain, wherein the data representation is relatively consistent (see \eg VGGNet \cite{Simonyan15} and ResNet \cite{He2016resnet}), methods for 3D point cloud representation are still developing. The most prevalent existing approaches can be roughly categorised as: point-based \cite{qi2017pointnet, qi2017pointnetplusplus}, voxel-based \cite{maturana2015voxnet, wu20153dshapenet, graham2018sparseconv, choy20194d}, and multiview-based \cite{su2015multiview, qi2016multiview, dai20183dmv}. 
PointNet \cite{qi2017pointnet} is one of the pioneering point-based approaches. It exploits a shareable multi-layer perceptron (MLP) network to extract per-point representation. PointNet++ \cite{qi2017pointnetplusplus} extends this approach by introducing a nested hierarchical PointNet to extract local context information. Although simple, PointNet and PointNet++ are still widely used in the tasks of semantic segmentation \cite{qi2017pointnetplusplus, xu2018spidercnn, wang2019graph}, instance segmentation \cite{wang2019asis, wang2018sgpn, he2020eccvembedding, he2020eccvmemory}, and 3D detection \cite{yang2019std, qi2019deep}.
Multi-view solutions often involve view projection to utilize well-explored 2D techniques. In~\cite{su2015multiview}, for instance, view-pooling is used to combine information from different views of a 3D shape and thereby to construct a compact shape descriptor. 3DSIS~\cite{hou20193dsis}, in contrast, projects features extracted from 2D views into 3D space. 
Voxel-based methods first transfer 3D points into rasterized voxels and apply convolution operations for feature extraction. Traditional 3D convolution methods \cite{maturana2015voxnet, song2016ssc} are often constrained by inefficient computation and limited GPU memory. In addition, computational and representational resources are wasted on void space. \SexyName, in contrast, uses sparse volumetric convolution \cite{graham2018sparseconv, choy20194d} to efficiently process this inherently sparse data. Focusing computation on the data, rather than the space it occupies, makes \SexyName faster, more robust, and better able to extract local patterns. 

\begin{figure*}[htbp]
	\begin{center}
		{
			\includegraphics[width=17.5cm]{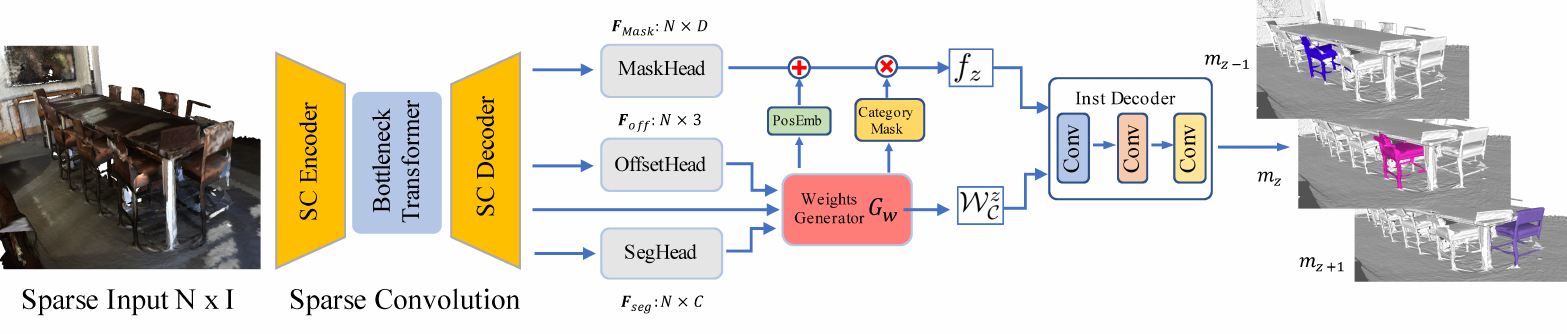}
			
		}
	\end{center}
\vspace{-0.3cm}
	\caption{
		\textbf{The structure of \SexyName}. It contains three main components: (1) a sparse convolution backbone based on~\cite{graham2018sparseconv}, which contains a light-weight transformer and outputs three parallel heads for instance mask generation, offset prediction, and semantic segmentation.  (2) A weight generator that takes centroid predictions and semantic segmentations as input. Homogenous points that have close votes for instance centroids and share the category predictions are explored to output instance-aware position embeddings, category-specific masks, and convolutional filters. (3) An instance decoder. Binary masks of instances are decoded by applying several convolutions, with the filters constructed by the Weight Generator. 
	}
	\label{fig:framework}
\end{figure*}

\textbf{Instance Segmentation of 3D Point Cloud.} 
As in the 2D image domain, 3D instance segmentation approaches can be broadly divided into two groups: top-down and bottom-up. Top-down methods often use a detect-then-segment approach, which first detects 3D bounding boxes of the instances and then predicts foreground points. 3D-BoNet \cite{yang20193dbonet}, for instance, first detects unoriented 3D bounding boxes from a single global representation by utilizing a Hungarian matching algorithm. Then per-point features are explored within each bounding box to mask out the background. Instead of regressing bounding boxes for instance proposals, GSPN \cite{yi2018gspn} generates instance shapes and applies analysis-by-synthesis.
Bottom-up methods, in contrast, group sub-components into instances. Methods applying this approach have dominated the leaderboard of the ScanNet dataset \cite{dai2017scannet}\footnote{\url{http://kaldir.vc.in.tum.de/scannet_benchmark/}}. The grouping techniques vary from simple clustering \cite{han2020occuseg, wang2019asis, he2020eccvmemory, phamjsis3dcvpr19, he2020eccvembedding, jiang2020pointgroup} to complex graph-based algorithms \cite{Engelmann20CVPR, han2020occuseg} based on learned embeddings. 
ASIS \cite{wang2019asis}, for example, learns point-level embeddings, regularized by a discriminative loss function \cite{bra2017cvprdis}, which encourages points belonging to the same instance to be mapped to similar locations in a metric space while separating points belonging to different instances. A mean-shift algorithm is then applied to generate instance masks. Many subsequent works \cite{he2020eccvembedding, phamjsis3dcvpr19, he2020eccvmemory, zhao2020jsnet} use the same general pipeline.
PointGroup \cite{jiang2020pointgroup} generates instance clusters from two sets of points: original and centroid shifted points. A network the authors label ScoreNet is used to evaluate the candidates. 
OccuSeg \cite{han2020occuseg}, in contrast, uses multi-task learning to generate feature embeddings, but also explicit occupancy embeddings that enable metric instance scale calculations to be made. 

\textbf{Dynamic Convolution.} The existing works most closely related to \SexyName are \cite{debrabandere16dynamic} and \cite{tian2020conditional}. Dynamic convolution was first proposed to enhance filter representation by encoding sample-specific and position-specific knowledge. 
CondInst \cite{tian2020conditional} successfully applies it in the 2D image domain for instance segmentation. However, our experiments demonstrate that it performs poorly when applied directly to 3D point clouds for the following reasons: 1) it introduces a large amount of computation, resulting in optimization difficulties. 2) the performance is constrained by the limited receptive field and representation capability due to the sparse convolution.
In this paper, we improve the dynamic convolution tailored for 3D point cloud instance segmentation and demonstrate its effectiveness and robustness on multiple benchmarks. Details
are 
introduced in the next section.

\section{Methods}

\subsection{Overall Architecture}
The structure of \SexyName is depicted in Fig.~\ref{fig:framework}. 
The input to the network is a matrix recording the point features $\textbf{P} \in \mathbb{R}^{N\times I}$, where $N$ is the total number of points and $I$ is the dimension of each point feature.
The goal is to predict a set of point-level binary masks and their corresponding category labels, denoted as $\{(\hat m_k, \hat c_k)\}$, where $\hat m_k \in \{0,1\}^N$, and $\hat c_k \in \{1, 2, \cdots, C\}$. $C$ is 20 for ScanNetV2 \cite{dai2017scannet} and 13 for S3DIS \cite{armeni2016s3dis}. 
Compared with previous top-performing approaches \cite{jiang2020pointgroup, Engelmann20CVPR}, where instances masks are dependent on the proposals, 
our method is proposal-free and can produce instance masks using only a small number of simple convolutional layers. 
The associated convolution filters are dynamically generated, conditioned on both spatial distribution of the data and the semantic predictions. As shown in Fig.~\ref{fig:framework}, \SexyName is comprised of three primary components:
(1) a backbone network, which is based on sparse convolution for feature extraction, and contains a light-weighted transformer \cite{Vaswani2017attention}, aiming to enlarge the receptive field and capture long-range dependencies. 
(2) A weight generator that responds to the individual characteristics of each instance to dynamically generate the appropriate filter parameters. To make the filters discriminative, a large category-specific context is introduced. 
(3) An instance decoder. Instances are separated in parallel, using only three convolution layers, by convolving the generated class-aware filters with position embedded features.

\subsection{Backbone Network}
\label{section:backbone}
Although our method is not restricted to any specific choice of backbones, we select sparse convolution \cite{graham2018sparseconv} for its efficiency and competitive performance. Following \cite{graham2018sparseconv, jiang2020pointgroup}, we construct a U-Net, which consists of an encoder and a decoder that have symmetrical structures. 
However, sparse convolution is often constrained by a limited receptive field and representation capability, due to the small number of convolution layers and channels. To this end, we propose a light-weight transformer \cite{Vaswani2017attention} to enhance long-range interactions on top of the encoder. 
The transformer is identical to the implementation of  \cite{Vaswani2017attention}, except for the position embedding layer, where the position-sensitive information is encoded as the mean of the pairwise direction vector. More details can be found in the supplementary materials. 

We denote the features output by the backbone as $\textbf{F}_b \in \mathbb{R}^{N \times D}$, where $D$ is the dimension of the output channel. Three parallel branches are built upon $\textbf{F}_b$ for semantic segmentation ($\textbf{F}_\text{seg} \in \mathbb{R} ^{N \times C}$), offset prediction ($\textbf{O}_\text{off} \in \mathbb{R}^{N \times 3}$), and instance masking ($\textbf{F}_\text{mask} \in \mathbb{R}^{N\times D}$), where $C$ is the category number. 

\textbf{Semantic Segmentation.} We apply traditional cross entropy loss $\mathcal{L}_\text{seg}$ for semantic segmentation. Pointwise prediction of the category label can be easily obtained, indicated as $\{l_\text{seg}^i\}_\text{i=1}^N$.

\textbf{Centroid Offset.} 
The variability in the distribution of points across surfaces makes aggregating contextual information complex.
To address the problem, we follow VoteNet \cite{qi2019deep}, by shifting points towards the corresponding centroids of instances. Point-wise prediction $o_\text{off}^i$ is supervised by the following loss function:
\begin{equation}
\mathcal{L}_\text{ctr} = \frac{1}{N_v}\sum_{i=0}^{N}\| p^i + o_\text{off}^i- ctr_\text{gt}^i\|\cdot \mathds{1}(p^i)
\end{equation} 
where $p^i$ is the coordinates of the $i$-th point, $o_\text{off}^i$ is the $i$-th item of $\textbf{O}_\text{off}$, and $ctr_\text{gt}^i$ is the geometric centroid of the corresponding instance. $\mathds{1}(p^i)$ is an indicator function, representing whether $p^i$ is a valid point for centroid prediction. $N_v$ is the total number of the valid points. For example, the categories of `floor' and `wall' are ignored for instance segmentation on ScanNetV2 \cite{dai2017scannet}, making them free from offset predictions. 
\iffalse 
\begin{figure}[t]
	\centering
	{
		\includegraphics[width=8.5cm]{figs/generator.pdf}
	}
	\caption{ The pipeline of the weight generator. Homogenous points are clustered by exploring both category prediction and geometric distribution. A light-weight sub-network is then applied to incorporate the larger context and applied once for each cluster to generate the convolution parameters used in instance decoding. Each filter is responsible for one instance.
	}
	\label{fig:generator}
\end{figure}
\fi 

\begin{figure*}[t]
	\centering
	{
		\includegraphics[width=.758\textwidth]{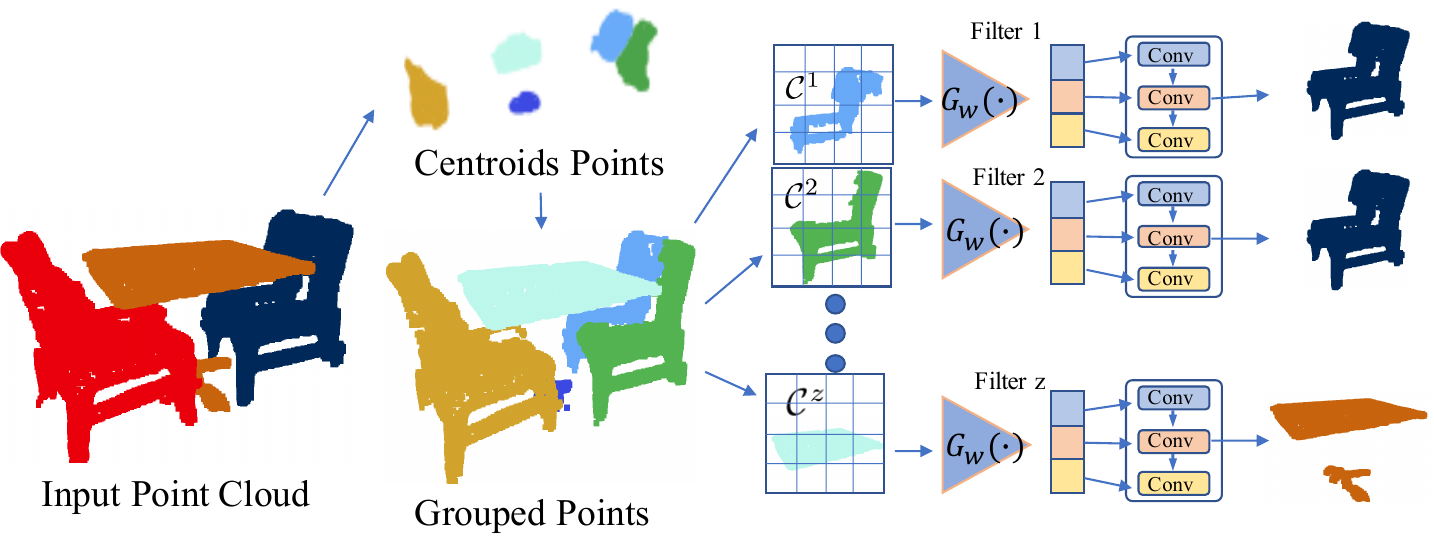}
	}
	\caption{ The pipeline of the weight generator. Homogenous points are clustered by exploring both category prediction and geometric distribution. A light-weight sub-network is then applied to incorporate the larger context and applied once for each cluster to generate the convolution parameters used in instance decoding. Each filter is responsible for one instance.
	}
	\label{fig:generator}
\end{figure*}

\subsection{Dynamic Weight Generator}
\label{section: filter}
The combination of the shallow network architecture and sparse convolution would typically cause a limited receptive field, and impair the method's ability to exploit large-scale context. To generate discriminative filters for distinguishing different instances we propose to group homogenous points that have close votes for the geometric centroids and share the category predictions. Then instance-aware filters are dynamically generated by applying a small sub-network for large context aggregation, as shown in Fig. \ref{fig:generator}.
Provided both predicted semantic labels and centroids offsets, we are ready for grouping homogenous points by using a similar strategy to that in \cite{jiang2020pointgroup}. Details of the algorithm can be found in the supplementary materials. However, different from \cite{jiang2020pointgroup} that directly treats the clusters as individual instance proposals, our method explores the spatial distribution of these points and integrates large context to generate filters for instances decoding. 
Due to the removal of the reliance on the quality of the instance proposals, the performance of our method is robust to the pre-defined hyper-parameters, as approved in the following experiments. Qualitative results are presented in Fig. \ref{fig:main}.
Moreover, compared with CondInst \cite{tian2020conditional}, where filters are generated for every valid pixel, \SexyName generates much less number of instance candidates (less than 60), and each filter is responsible for one instance in a specific class, reducing the difficulties for optimization and the heavy requirements for hardware resources. 

Given point-wise offset prediction $\{o_\text{off}^i \} _\text{i=1}^N$, centroids distribution $\{p_\text{ctr}^i \in \mathbb{R}^3 \}_\text{i=1}^N$ can be easily calculated by $p_\text{ctr}^i=p^i + o_\text{off}^i$. With $\{p_\text{ctr}^i\}_{i=1}^N$ and semantic labels $\{l_\text{seg}^i \} _{i=1} ^N$, instances are separated to a certain extent. We explore the void spaces among instances by applying a breadth-first searching algorithm \cite{jiang2020pointgroup} to group homogenous points that have identical semantic labels and close centroids predictions.
Point $p^j$ can be grouped with $p^i$ if it satisfies:
(1) $l_\text{seg}^j = l_\text{seg}^i$. (2) $\|p_\text{ctr}^j - p_\text{ctr}^i\|_2 <= r$, where $r$ is a pre-defined searching radius. The whole algorithm can be found in the supplementary materials. The grouping process ends up with a set of clusters $\{\mathcal{C}^z\}_\text{z=1}^Z$, where $Z$ refers to the total number of clusters. 
As only one specific category is considered for each cluster, semantic label $l_\mathcal{C}^z \in \mathbb{R}$ of cluster $\mathcal{C}^z$ can be easily obtained from the semantic prediction. We also label the geometrical centroids for cluster $\mathcal{C}^z$ as $\mathcal{C}_\text{ctr}^z \in \mathbb{R} ^3$, which is calculated as the average of the coordinates of the points in $\mathcal{C}^z$.
Each cluster contains a bunch of points that are distributed across the instance, introducing a large context and rich geometric information. We explore the clusters and generate instance-aware weights for responding to the individual characteristics of each instance. 

For cluster $\mathcal{C}^z$, we first voxelize it with a grid size of $g$, which is set to 14 in all our experiments. The features of each grid is calculated as the average of the point feature $\textbf{F}_b$ within the grid, where $\textbf{F}_b$ is the output of the backbone. 
To aggregate context for cluster $\mathcal{C}^z$, a light-weighted sub-network $G_w(\cdot)$ is maintained. It contains two sparse convolutional layers with a kernel size of 3, a global pooling layer, and an MLP layer. The output is all convolutional parameters flattened in a compact vector, $\mathcal{W}_\mathcal{C}^z$. Each $\mathcal{W}_\mathcal{C}^z$ is responsible for one specific instance.
The size of $\mathcal{W}_\mathcal{C}^z$ is decided by the feature dimension and the number of the subsequent convolution layers (see Eq. \eqref{eq:filter_calcu}).

\subsection{Instance Decoder}
\label{section: instance_gen}
Given a specific category, position representation is critical to separate different instances. To encode position sensitive knowledge, we directly append position embeddings in the feature space. For $z$-th instance with geometric centroid of $\mathcal{C}_\text{ctr}^z$, position embedding for the $i$-th point $f_\text{pos}^i$ is calculated as:
\begin{equation}
f_\text{pos}^i = p^i - \mathcal{C}_\text{ctr}^z
\end{equation}
where $p^i$ is the coordinates of the $i$-th point. For $\mathcal{W}_\mathcal{C}^z$, the input feature $\{f_z^i \in \mathbb{R}^{D+3} \}_{i=1}^{N} $ is generated by concatenating $\{f_\text{pos}^i \in \mathbb{R}^3 \}_{i=1}^N$ and $\{f_\text{mask}^i \in \mathbb{R}^D \}_{i=1}^N $.
Provided both instance-aware filters $\{\mathcal{W}_\mathcal{C}^z\}_{z=1}^Z$ and position-embedded features $\{f_z \in \mathbb{R}^{N\times (D+3)} \}_{z=1}^Z$, we are ready to decode binary segmentations of instances. The whole decoder contains three convolution layers with a kernel size of $1\times1$. Each layer uses ReLU as the activation function without normalization. Supposing the feature dimension of $f_\text{mask}^i$ is 8, meaning $D=8$, and the feature dimension of the decoder is 8, the total number of parameters (including both weights and biases) of $\mathcal{W}_\mathcal{C}^z$ is 177, which is calculated by:
\begin{equation}
177 = \underbrace {(8+3) \times 8 + 8}_{conv1} + \underbrace{8 \times 8 + 8}_{conv2} + \underbrace{ 8 \times 1 + 1}_{conv3}
\label{eq:filter_calcu}
\end{equation}
Formally, the instance decoder is formulated as:
\begin{equation}
m_z = {\rm Conv}(\mathcal{W}_\mathcal{C}^z, f_z)
\end{equation}
where $m_z \in \mathbb{R}^N$ is the predicted binary mask for the $z$-th instance. 
Also, as filters are derived from a set of points that have identical semantic labels, we propose to operate the convolution on the points that have the same semantic predictions with $l_\mathcal{C}^z$.
During training, the ground truth for $\mathcal{C}^z$ is $\hat m_z$ if it has the largest number of points in $\mathcal{C}^z$. 
The loss function for instance segmentation is defined as:
\begin{equation}
\mathcal{L}_\text{mask} = \frac{1}{Z} \sum_{z=1}^{Z} \frac{1}{N_z} \sum_{j=1}^{N} \mathds{1}_{l_\text{seg}^j=l_\mathcal{C}^z} \cdot L_\text{BCE}(m_z^j, \hat m_z^j)
\label{eq:conv}
\end{equation}
where $Z$ is the total number of the clusters, $l_\text{seg}^j$ is the semantic prediction of the $j$-th point, $l_\mathcal{C}^z$ is the semantic label of the $z$-th cluster, and $L_\text{BCE}$ is the binary cross entropy loss function.
$\mathds{1}$ is an indicator function, showing the loss is only computed on the points that have identical semantic labels with group $\mathcal{C}^z$, and $N_z$ is a normalization item which is calculated as: $\sum_{j=1}^N \mathds{1}_{l_\text{seg}^j=l_\mathcal{C}^z}$.
In addition to the point-wise supervision, we also utilize the dice loss \cite{tian2020conditional} $\mathcal{L}_\text{dice}$, which is designed for addressing the imbalance between the foreground and background points. 

\subsection{Training details}
The loss function of \SexyName can be formulated as:
\begin{equation}
\mathcal{L} = \mathcal{L}_\text{seg} + \mathcal{L}_\text{ctr} +  \mathcal{L}_\text{mask} + \mathcal{L}_\text{dice}
\end{equation}
where $\mathcal{L}_\text{seg}$ is for semantic segmentation, 
$\mathcal{L}_\text{ctr}$ is for instance centroids supervision, and
$\mathcal{L}_\text{mask}$ and $\mathcal{L}_\text{dice}$ are two loss items for instance segmentation. All loss weights are set to 1.0. 

During the inference time, we perform NMS on the instance binary masks $ \{m_\mathcal{C}^z\}_{z=1}^Z$, which are scored by the mean value of the semantic scores among the foreground points. The IoU threshold is the same as \cite{jiang2020pointgroup}, with a value of 0.3. Cluster $\mathcal{C}^z$ is ignored if it contains points less than 50.
The voxel size is set to 0.02m and 0.05m for ScanNetV2 \cite{dai2017scannet} and S3DIS \cite{armeni2016s3dis}, respectively.
The hyper-parameter $r$ of the searching radius is set to 0.03m, which is the same with \cite{jiang2020pointgroup} for a fair comparison.
We implement multi-GPU training with a batch size of 16 and 4 GPUs. For the first 6k iterations, we only train the semantic segmentation $\mathcal{L}_\text{seg}$ and centroid prediction $\mathcal{L}_\text{ctr}$, as dynamic filters depend on the results of both tasks. For the next 24k iterations, we compute all the loss items. During the training, the initial learning rate is set to 0.01 with an Adam optimizer.
We apply the same data augmentation strategy with \cite{jiang2020pointgroup}, including random cropping, flipping, and rotating.

\section{Experiments}

To validate the effectiveness of our proposed method, we conduct both qualitative and quantitative experiments on datasets that are widely studied: ScanNetV2 \cite{dai2017scannet} and Stanford 3D Indoor Semantic Dataset (S3DIS) \cite{armeni2016s3dis}. In this section, we show that our method demonstrates superiority in both effectiveness and efficiency.

\subsection{Datasets}
S3DIS contains 13 categories that commonly exist in indoor scenes. The point cloud data is collected in 6 large-scale areas, covering more than 6000 $m^2$ with more than 215 million points. Following the protocols of previous methods \cite{wang2018sgpn, he2020eccvmemory}, we evaluate the performance on Area-5 and train the model on the other sets. ScanNet \cite{dai2017scannet} is another large-scale benchmark for indoor scene analysis, which consists of 1613 scans with 40 categories in total. The dataset is split into 1201, 312, and 100 for training, evaluating, and testing, respectively. 
Like previous methods, we estimate the performance of instance segmentation on 18 common categories.
Also, we follow the strategy in 3D-MAP \cite{Engelmann20CVPR} and report the performance of 3D detection, where the results are obtained by fitting an axis-aligned bounding box around the instance segmentation.  

\begin{table}[!t]
	\small 
	\centering
	\addtolength{\tabcolsep}{-3.0pt}
	
	\begin{center}
		\begin{tabular}{c|cccc|c|c|c}
			\toprule[0.12 em]
			
			Method & Group  & PosEmb &CAD  &TF & mAP &AP@50 &AP@25  \\
			\toprule[0.12 em]
			Baseline & & & & &24.8 &43.8 &56.4 \\
			CondInst & & \checkmark& & &27.0 &44.7 &57.5 \\
			
			&\checkmark & & & &29.4 &49.7 &66.3\\
			&\checkmark &\checkmark & & &31.8&52.9 &68.4\\
			&\checkmark &\checkmark &\checkmark & &34.1 &55.3 &69.5\\
			Ours&\checkmark &\checkmark &\checkmark &\checkmark &35.4 &57.6 &72.9\\			
			\bottomrule[0.1 em]
		\end{tabular}
	\end{center}
	\caption{Ablation studies on the components of our proposed method. We evaluate the performance on the ScanNetV2 \cite{dai2017scannet} validation set. $\textbf{Group}$ indicates that the dynamic filters are generated by gathering homogenous points that share the semantic labels and have close centroids votes. \textbf{PosEmb} refers to the position embeddings $f_\text{pos}$. $\textbf{CAD}$ denotes the category-aware decoding that the convolution in the decoding process is only operated on category-specific points, instead of all points. $\textbf{TF}$ refers to the light-weight transformer applied for the backbone.
	}
	
	\label{tab:ablation_study}
\end{table}

\subsection{Evaluation Metrics}
For ScanNetV2, we report the metric of mean average precision (mAP), which is widely used in the 2D image domain. AP@50 and AP@25 are also provided, having an IoU threshold set to 0.5 and 0.25, respectively.
For S3DIS, we apply the metrics that are used in \cite{wang2019asis, he2020eccvembedding, he2020eccvmemory}: mConv, mWConv, mPrec, and mRec. 
mConv is defined as the mean instance-wise IoU. mWConv denotes the weighted version of mConv, where the weights are determined by the sizes of instances. mPrec and mRec denote the mean precision and recall, respectively. 

\subsection{Ablation Studies}
In this section, we analyze the effect of each component in our proposed \SexyName. Performance is reported in terms of mAP, AP@50, and AP@25. All experiments are conducted with the same setting and training schedule, and are evaluated on ScanNetV2 \cite{dai2017scannet} validation set.

\textbf{Baseline.} 
We build a strong baseline by generating filters for each foreground point without introducing any clustering operation. Due to the large size of $N$, we randomly select 150 points for instance decoding. As presented in Table \ref{tab:ablation_study}, our method achieves 24.8, 43.8, and 56.4 in terms of mAP, AP@50, and AP@25, respectively. We also implement CondInst \cite{tian2020conditional}, which has demonstrated its success in the 2D image domain. As presented in the second row in Table \ref{tab:ablation_study} the mAP has boosted by 2.2\%, with the help of instance-related position embeddings. 

\textbf{Ablation on the Grouping Homogenous Points.} Due to the limited receptive field introduced by the sparse convolution, it is significant to incorporate rich context for distinguishing different instances. To this end, we propose to integrate homogenous points that are defined in Sec. \ref{section: filter}. Thanks to the grouping operation, the model surpasses the baseline by a large margin in terms of all metrics. Besides, the grouping operation reduces the number of instance candidates (less than 60), lowering the optimization difficulties and the heavy requirements for the hardware facilities.

\begin{figure}
	\centering
	{
		\includegraphics[width=6cm]{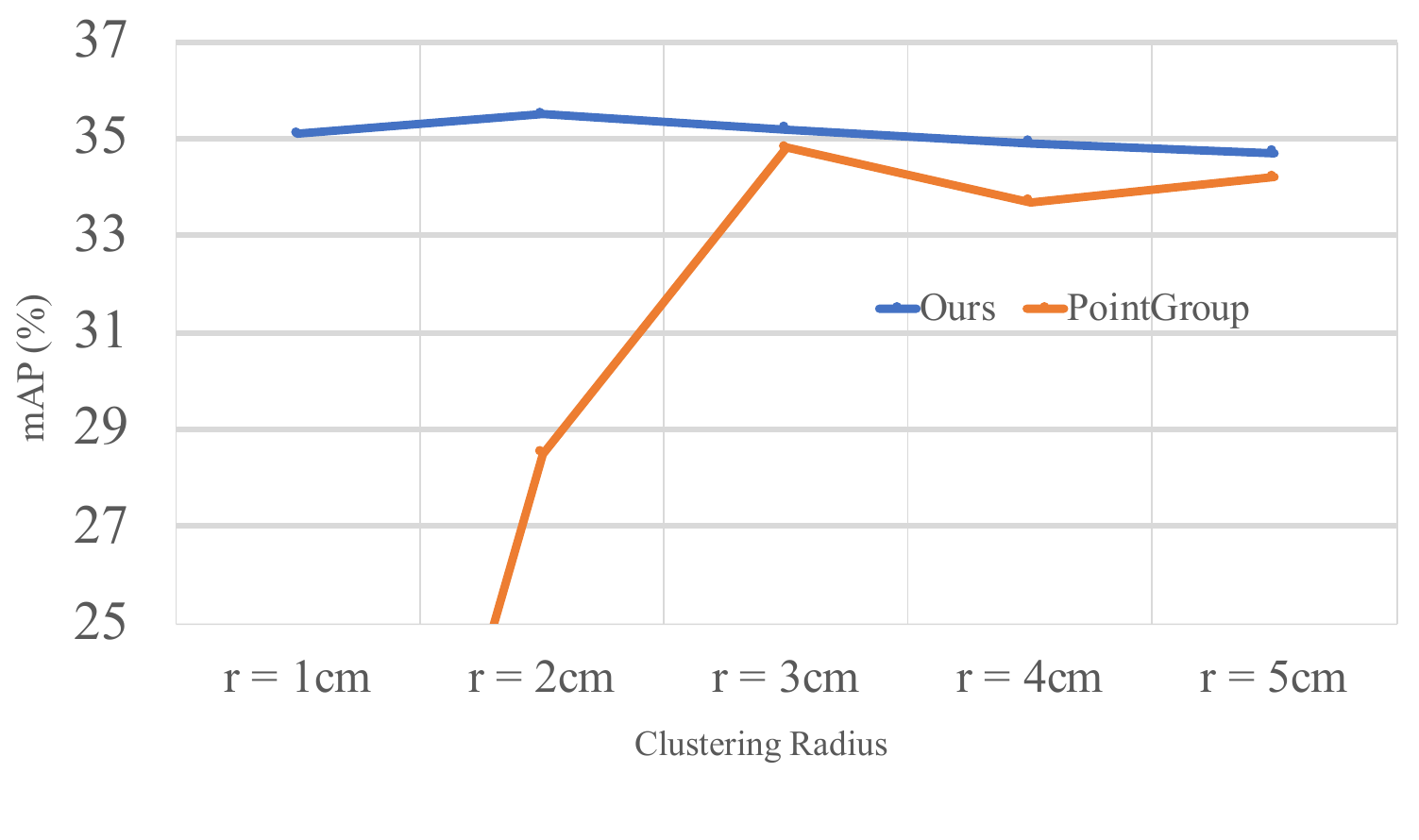}
	}
	\vspace{-0.2cm}
	\caption{The performance of the instance segmentation with different clustering radius $r$. All numbers for PointGroup are obtained from the paper or tested by the provided model. Unlike PointGroup, which is sensitive to the hyper-parameter and requires heuristic tuning, our method shows strong robustness.
	}
	\label{fig:radius}
\end{figure}
\begin{figure*}[tbp]
	\centering
	{
		\includegraphics[width=15cm]{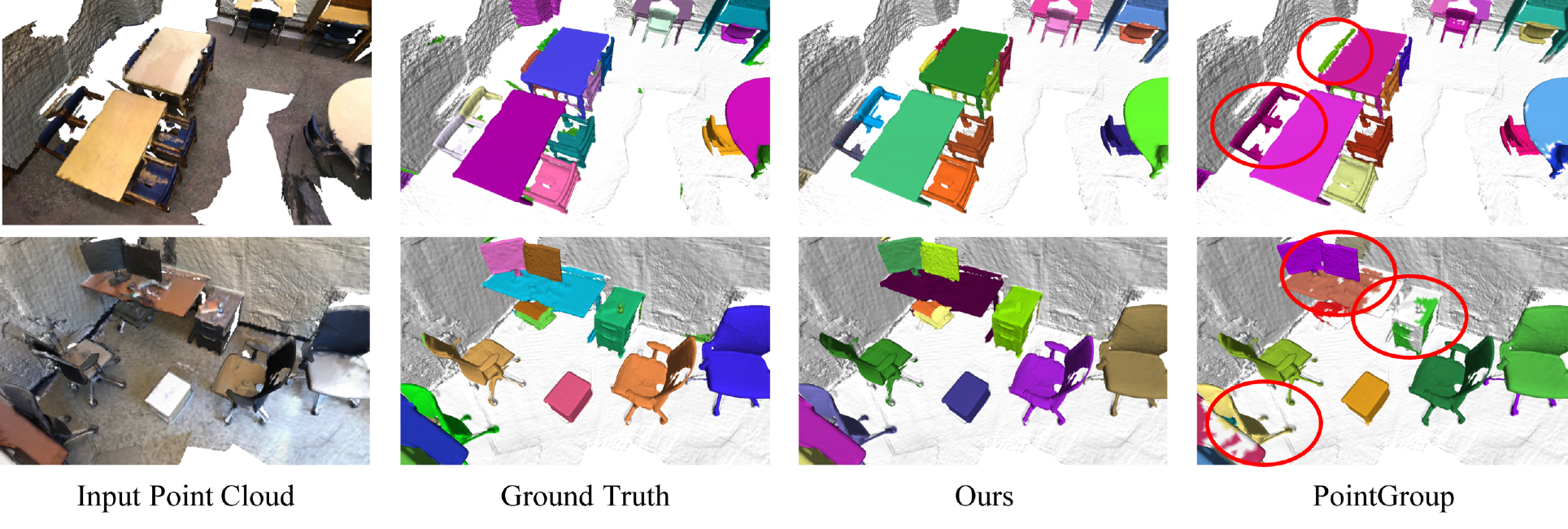}
	}

	\caption{Comparison of the results with PointGroup \cite{jiang2020pointgroup}. The ellipses highlight specific over-segmentation/joint errors. 
	}
	\label{fig:compg}
\end{figure*}

\textbf{Ablation on the Category-Aware Decoding.} As filters are generated by exploring the points that have identical semantic predictions, only certain category context is encoded. We propose to convolve each filter on these category-specific points and mask out irrelevant points. As presented in Table \ref{tab:ablation_study}, adding category masks improves the mAP from 31.8\% to 34.1\%. 

\textbf{Ablation on the Transformer.} 
As limited receptive field and representation ability introduced by the sparse convolution, we propose to add a light-weighted transformer upon the bottleneck layer to capture the long-range dependencies and enhance interactions among different points, while maintaining efficiency. As presented in Table \ref{tab:ablation_study}, the transformer brings about 1.3\% improvements in terms of mAP.

\textbf{Ablation on the Clustering Radius.} 
The clustering radius is pre-defined in the grouping step. PointGroup \cite{jiang2020pointgroup}, which treats clusters as the instance proposals, makes the performance highly dependent on the quality of the clustering results. 
We test the performance with a different radius, as shown in Fig. \ref{fig:radius}. Grouping with a small $r$ may generate over-segmented results, while a large $r$ increases the risk of merging two adjacent objects. 
As a result, changing the radius $r$ from 3cm to 2cm drops mAP by 6.3\%, and 23.9\% by changing $r$ from 3cm to 1cm. The volatility makes it necessary to be carefully tuned, demonstrating limited generalization capability to various scenes. Our method, on the other hand, eliminates the dependence on the proposals, showing strong robustness to the radius $r$. More qualitative results can be found in Fig. \ref{fig:compg}.

\textbf{Analysis on Efficiency.} 
Different from previous point-based approaches that require to split each scene as 1m$\times$1m blocks and apply a complex block merging algorithm \cite{wang2018sgpn, he2020eccvembedding, he2020eccvmemory, wang2019asis}, our method takes the whole scene as input. 
In addition, we also compare our \SexyName with PointGroup, which has shown its efficiency on large-scale scenes. We report the inference time that is averaged on the whole validation set. With the only post-processing step NMS, our method runs at 0.28s per scan on a 1080TI GPU, while the PointGroup runs at 0.39s with the same facility. 

\subsection{Comparison with State-of-the-art Methods}
\textbf{Object Detection.} Following \cite{Engelmann20CVPR}, we report the performance of 3D object detection on the validation set of ScanNetV2, which is obtained by fitting axis-aligned bounding boxes containing the instances.
As shown in Table \ref{tab:object_detection}, our method surpasses PointGroup \cite{jiang2020pointgroup} by 3.1\% and 3.0\% in terms of AP@25 and AP@50, respectively, demonstrating the compactness of our generated instance masks.

\begin{table}[!t]
	\small 
	\centering
	\addtolength{\tabcolsep}{-3.0pt}
	
	\begin{center}
		\begin{tabular}{r  c c}
			\toprule[0.12 em]
					
					\multicolumn{3}{c}{\textbf{3D Object Detection}}  \\ 
					\midrule
					ScanNetV2     & AP@25\% & AP@50\%  \\
					\midrule
					DSS~\cite{Song2016cvpr}      & 15.2  & 6.8  \\
					MRCNN 2D-3D~\cite{he2018maskrcnn}& 17.3  & 10.5 \\
					F-PointNet~\cite{qi2017frustum} & 19.8  & 10.8 \\
					GSPN~\cite{yi2018gspn}       & 30.6  & 17.7 \\
					3D-SIS~\cite{hou20193dsis}    & 40.2  & 22.5 \\
					VoteNet~\cite{qi2019deep}    & 58.6  & 33.5 \\
					
					PointGroup~\cite{jiang2020pointgroup} &56.8 &42.3 \\
					\textbf{Ours}                 &\textbf{58.9}  &\textbf{45.3} \\
					
			\bottomrule[0.1 em]
		\end{tabular}
	\end{center}
\vspace{-0.1cm}
	\caption{3D object detection on the validation set of ScanNetV2~\cite{dai2017scannet}. We report per-class average precision (AP) with IoU thresholds of 25 \% and 50 \%. The performance of PointGroup \cite{jiang2020pointgroup} is evaluated with the provided model. We use the same backbone as in \cite{jiang2020pointgroup} for a fair comparison.
	}
	
	\label{tab:object_detection}
\end{table}

\begin{table}[!t]
	\small 
	\begin{center}
		\begin{tabular}{ r |c|c|c|c}
			\toprule
						Method  & mCov    & mWCov    &  mPrec & mRec \\

						\hline
						SGPN'18 ~\cite{wang2018sgpn}   &  32.7  & 35.5  &  36.0  & 28.7   \\
						
						ASIS'19 ~\cite{wang2019asis}   & 44.6  &  47.8 &  55.3 &  42.4 \\
						3D-BoNet'19 ~\cite{yang20193dbonet}   & -  &  - &  57.5 &  40.2 \\
						3D-MPA'20~\cite{Engelmann20CVPR}   & - &- &63.1 &58.0\\
						MPNet'20~\cite{he2020eccvmemory}      & {50.1} &{53.2} &{62.5} &{49.0} \\
						
						InsEmb'20~\cite{he2020eccvembedding}   &49.9 & 53.2 & 61.3 &48.5 \\
						PointGroup'20~\cite{jiang2020pointgroup}  &- &- & 61.9 & 62.1 \\
						\textbf{Ours}  &\textbf{63.5} &\textbf{64.6} &\textbf{64.3} &\textbf{64.2} \\
						
			\bottomrule
		\end{tabular}
	\end{center}

	\caption{The results of instance segmentation on the S3DIS dataset. Performance on Area-5 is reported. A comparison with previous top-performing approaches is presented.
	}
	
	\label{tab:s3dis_ins_results}
\end{table}

\begin{table*}
	\resizebox{\textwidth}{!}{
		\begin{tabular}{ r |cc|cccccccccccccccccc}
			\toprule
			&\textbf{AP@50} &\textbf{mAP}& \rotatebox{90}{cabinet} & \rotatebox{90}{bed} & \rotatebox{90}{chair} & \rotatebox{90}{sofa} & \rotatebox{90}{table} & \rotatebox{90}{door} & \rotatebox{90}{window} & \rotatebox{90}{bookshe.} & \rotatebox{90}{picture} & \rotatebox{90}{counter} & \rotatebox{90}{desk} & \rotatebox{90}{curtain} & \rotatebox{90}{fridge} & \rotatebox{90}{s.curtain} & \rotatebox{90}{toilet} & \rotatebox{90}{sink} & \rotatebox{90}{bath} & \rotatebox{90}{otherfu.} \\
			\midrule
			SegClu~\cite{hou20193dsis}&10.8 &- &10.4&11.9&15.5&12.8&12.4&10.1&10.1&10.3&0.0&11.7&10.4&11.4&0.0&13.9&17.2&11.5&14.2&10.5\\
			MRCNN~\cite{he2018maskrcnn}&9.1 &- &11.2&10.6&10.6&11.4&10.8&10.3&0.0&0.0&11.1&10.1&0.0&10.0&12.8&0.0&18.9&13.1&11.8&11.6\\
			SGPN~\cite{wang2018sgpn}&11.3 &- &10.1&16.4&20.2&20.7&14.7&11.1&11.1&0.0&0.0&10.0&10.3&12.8&0.0&0.0&48.7&16.5&0.0&0.0\\
			3D-SIS~\cite{hou20193dsis}&18.7&- &19.7&37.7&40.5&31.9&15.9&18.1&0.0&11.0&0.0&0.0&10.5&11.1&18.5&24.0&45.8&15.8&23.5&12.9\\
			MPNet~\cite{he2020eccvmemory}&31.0&- &-&-&-&-&-&-&-&-&-&-&-&-&-&-&-&-&-&-\\
			MTML~\cite{Jean2019mtml}&40.2&- &14.5&54.0&79.2&48.8&42.7&32.4&32.7&21.9&10.9&0.8&14.2&39.9&42.1&64.3&96.5&36.4&70.8&21.5\\
			3D-MPA~\cite{Engelmann20CVPR}&{59.1} &35.3 &{51.9}&{72.2}&{83.8}&\textbf{66.8}&{63.0}&\textbf{43.0}&{44.5}&{58.4}&{38.8}&{31.1}&{43.2}&
			\textbf{47.7}&\textbf{61.4}&\textbf{80.6}&\textbf{99.2}&{50.6}&\textbf{87.1}&{40.3}\\
			PointGroup~\cite{jiang2020pointgroup} &56.9 &34.8 &48.1 &69.6 &87.7 &71.5 &62.9 &42.0 &46.2 &54.9 & 37.7 &22.4 &41.6 & 44.9&37.2 &64.4 &98.3 &61.1 &80.5 &53.0 \\
			\textbf{Ours-S}&57.6 &35.4 &50.6 &\textbf{73.8} &84.4 & 72.1 &\textbf{69.9} &40.8 &44.5 &\textbf{62.4} & 34.8 &21.2 & 42.2 &37.0 & 41.6 & 62.7 &92.9 &61.6 & 82.6 &47.5 \\
			
			\textbf{Ours-L} & \textbf{61.0} & \textbf{40.6} &\textbf{52.3} & 70.4 &\textbf{90.2} & 65.8 &69.6 & 40.5 & \textbf{47.2} &48.4 &\textbf{44.7} & \textbf{34.9} & \textbf{52.3} &47.5 &51.5 & 70.3 & 94.8 & \textbf{74.3} & 77.4 &\textbf{56.4} \\
			\bottomrule
		\end{tabular}
	}
	\caption{Per class 3D instance segmentation on ScanNetV2~\cite{dai2017scannet} validation set. Both mAP and AP@50 are reported.}
	\label{tab:scannet_val}
\end{table*}

\begin{figure*}[htbp]
	\centering
	{
		\includegraphics[width=15cm,height=7.1cm]{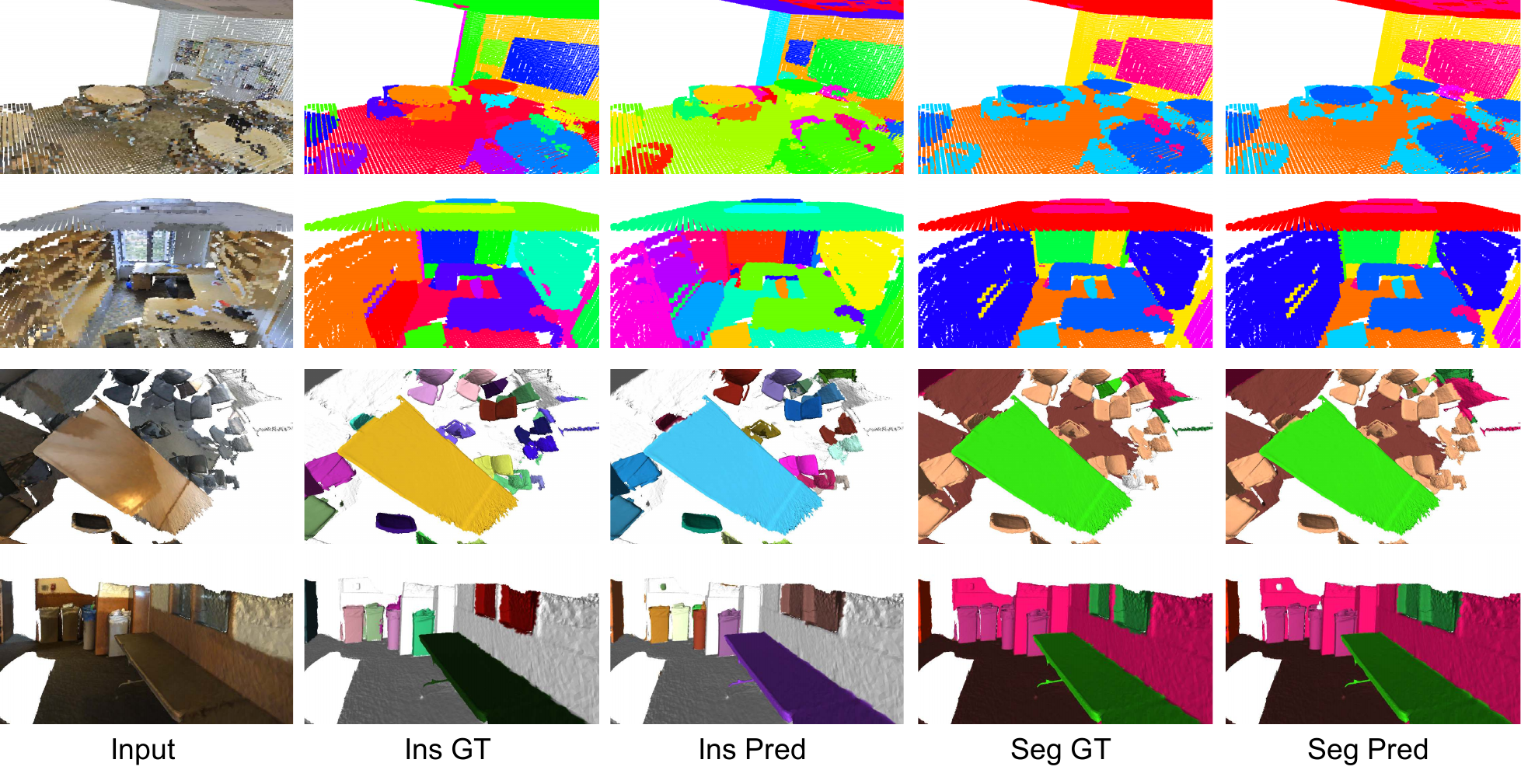}
	}
\vspace{-0.2cm}
	\caption{Visualization of semantic and instance segmentation results on both S3DIS (top) and ScanNetv2 (bottom) benchmarks. Instances are presented with random colors.}
	\label{fig:results}
\end{figure*}

\textbf{Instance Segmentation on S3DIS.}
We report the performance of instance segmentation on the S3DIS benchmark, as shown in Table \ref{tab:s3dis_ins_results}. 
Our method achieves the highest performance with all the evaluation metrics.
The results in terms of mPrec and mRec are 2.6\% and 2.1\% higher than PointGroup \cite{jiang2020pointgroup}. Our method also reaches 60.9\% under the metric of AP@50, which is 3.1\% higher than PointGroup \cite{jiang2020pointgroup}. We compute all these metrics with the evaluation code provided by \cite{wang2019asis}. Qualitative results are illustrated in Fig. \ref{fig:results}.

\textbf{Instance Segmentation on ScanNetV2.}
We report the results of instance segmentation on the validation and testing sets of ScanNetV2, as presented in Table \ref{tab:scannet_val} and Table \ref{tab:scannet_test}, respectively.
We report both AP@50 and mAP on the validation set. We implement \SexyName with both small and large backbones, denoted as $\textbf{Ours-S}$ and $\textbf{Ours-L}$, respectively. Two models share the same network structure but with a different number of channels for convolution. We first compare $\textbf{Ours-S}$ and PointGroup, which are implemented with the same backbone. Our method surpasses it by 0.7\% and 0.6\% in terms of AP@50 and mAP, respectively. We also make a fair comparison with 3D-MPA \cite{Engelmann20CVPR}, our large model surpasses it by 2.3\% and 4.9\% in terms of AP@50 and mAP, respectively.
We also report the performance of \SexyName on the test set, as shown in Table \ref{tab:scannet_test}. Highest AP@50 is achieved.

\begin{table*}[th!]
	\centering
	\resizebox{\textwidth}{!}{
		\begin{tabular}{ r |c|cccccccccccccccccc}
			\toprule[1pt]
			Method &  \textbf{AP@50} &\rotatebox{90}{bathtub}&\rotatebox{90}{bed}&\rotatebox{90}{bookshe.}&\rotatebox{90}{cabinet}&\rotatebox{90}{chair}&\rotatebox{90}{counter}&\rotatebox{90}{curtain}&\rotatebox{90}{desk}&\rotatebox{90}{door}&\rotatebox{90}{otherfu.}&\rotatebox{90}{picture}&\rotatebox{90}{refrige.}&\rotatebox{90}{s. curtain}&\rotatebox{90}{sink}&\rotatebox{90}{sofa}&\rotatebox{90}{table}&\rotatebox{90}{toilet}&\rotatebox{90}{window}\\
			\hline
			SGPN~\cite{wang2018sgpn} & 0.143 & 0.208 & 0.390 & 0.169 & 0.065 & 0.275 & 0.029 & 0.069 & 0.000 & 0.087 & 0.043 & 0.014 & 0.027 & 0.000 & 0.112 & 0.351 & 0.168 & 0.438 & 0.138 \\
			3D-BEVIS~\cite{elich20193dbevis} & 0.248 & 0.667 & 0.566 & 0.076 & 0.035 & 0.394 & 0.027 & 0.035 & 0.098 & 0.099 & 0.030 & 0.025 & 0.098 & 0.375 & 0.126 & 0.604 & 0.181 & 0.854 & 0.171 \\
			R-PointNet~\cite{yi2018gspn} &0.306 & 0.500 & 0.405 & 0.311 & 0.348 & 0.589 & 0.054 & 0.068 & 0.126 & 0.283 & 0.290 & 0.028 & 0.219 & 0.214 & 0.331 & 0.396 & 0.275 & 0.821 & 0.245 \\
			DPC~\cite{Engelmann20ICRA} & 0.355 & 0.500 & 0.517 & 0.467 & 0.228 & 0.422 & {0.133} & 0.405 & 0.111 & 0.205 & 0.241 & 0.075 & 0.233 & 0.306 & 0.445 & 0.439 & 0.457 & 0.974 & 0.23 \\
			3D-SIS~\cite{hou20193dsis} & 0.382 & 1.000 & 0.432 & 0.245 & 0.190 & 0.577 & 0.013 & 0.263 & 0.033 & 0.320 & 0.240 & 0.075 & 0.422 & 0.857 & 0.117 & 0.699 & 0.271 & 0.883 & 0.235 \\
			MASC~\cite{Liu2019masc} & 0.447 & 0.528 & 0.555 & 0.381 & 0.382 & 0.633 & 0.002 & 0.509 & 0.260 & 0.361 & 0.432 & 0.327 & 0.451 & 0.571 & 0.367 & 0.639 & 0.386 & 0.980 & 0.276 \\
			PanopticFusion~\cite{Narita2019iros} & 0.478 & 0.667 & 0.712 & 0.595 & 0.259 & 0.550 & 0.000 & 0.613 & 0.175 & 0.250 & 0.434 & 0.437 & 0.411 & 0.857 & 0.485 & 0.591 & 0.267 & 0.944 & 0.35 \\
			3D-BoNet~\cite{yang20193dbonet} & 0.488 & 1.000 & 0.672 & 0.590 & 0.301 & 0.484 & 0.098 & 0.620 & 0.306 & 0.341 & 0.259 & 0.125 & 0.434 & 0.796 & 0.402 & 0.499 & 	0.513 & 0.909 & 0.439 \\
			MTML~\cite{Jean2019mtml} & 0.549 & 1.000 & {0.807} & 0.588 & 0.327 & 0.647 & 0.004 & {0.815} & 0.180 & 0.418 & 0.364 & 0.182 & 0.445 & 1.000 & 0.442 & 0.688 & {0.571} & {1.000} & 0.396 \\
			PointGroup~\cite{jiang2020pointgroup} & {0.636} & {1.000} & 0.765 & {0.624} & {0.505} & {0.797} & 0.116 & 0.696 & {0.384} & {0.441} & {0.559} & {0.476} & {0.596} & {1.000} & {0.666} & {0.756} & 0.556 & 0.997 & {0.513} \\
			3D-MPA~\cite{Engelmann20CVPR} & {0.611} &1.000 &0.833  &0.765  &0.526  &0.756  &0.136  &0.588  &0.470  &0.438  &0.432  &0.358  &0.650  &0.857  &0.429  &0.765  &0.557  &1.000  &0.430  \\
			\textbf{Ours}&{0.641}  &1.000 &0.841  &0.893  &0.531  &0.802  &0.115  &0.588  &0.448  &0.438  &0.537  &0.430  &0.550  &0.857  &0.534  &0.764  &0.657  &0.987  &0.568  \\
			\bottomrule[1pt]
		\end{tabular}
	}

	\caption{3D instance segmentation results on ScanNetV2 testing set with AP@50 scores on 18 categories. 
	}

	\label{tab:scannet_test}
\end{table*}

\section{Conclusion}
Achieving robustness to the inevitable variation in the data has been one of the ongoing challenges in 3D point cloud segmentation.  We have shown here that dynamic convolution offers a mechanism by which to have the segmentation method actively respond to the characteristics of the data at test time, and that this does in-fact improve robustness.  It also allows devising an approach that avoids many other pitfalls associated with bottom-up methods. The particular dynamic-convolution-based method that we have proposed, \SexyName, not only achieves state-of-the-art results, it offers improved efficiency over existing methods.

{\small
\bibliographystyle{ieee_fullname}
\bibliography{pointcloud3d}
}

\end{document}